\documentclass[sigconf]{acmart}
\usepackage{caption}
\usepackage{subcaption}
\usepackage{tabularx}
\usepackage{float}
\AtBeginDocument{%
  }

\begin{document}

\title{Assessing Adversarial Robustness of Large Language Models: An Empirical Study}

\author{Zeyu Yang}
\email{zeyu.yang@telepathy.ai}
\affiliation{%
  \institution{Telepathy Labs}
  \city{Zürich}
  \country{Switzerland}
}
\author{Xiaochen Zheng}
\email{xzheng@ethz.ch}
\affiliation{%
  \institution{ETH Zürich}
  \city{Zürich}
  \country{Switzerland}
}
\author{Zhao Meng}
\email{zhmeng@ethz.ch}
\affiliation{%
  \institution{ETH Zürich}
  \city{Zürich}
  \country{Switzerland}
}
\author{Roger Wattenhofer}
\email{wattenhofer@ethz.ch}
\affiliation{%
  \institution{ETH Zürich}
  \city{Zürich}
  \country{Switzerland}
}


\begin{abstract}
Large Language Models (LLMs) have revolutionized natural language processing, but their robustness against adversarial attacks remains a critical concern. We presents a novel white-box style attack approach that exposes vulnerabilities in leading open-source LLMs, including Llama, OPT, and T5. We assess the impact of model size, structure, and fine-tuning strategies on their resistance to adversarial perturbations. Our comprehensive evaluation across five diverse text classification tasks establishes a new benchmark for LLM robustness. The findings of this study have far-reaching implications for the reliable deployment of LLMs in real-world applications and contribute to the advancement of trustworthy AI systems.
\end{abstract}


\maketitle

\section{Introduction}
In recent years, the field of artificial intelligence has witnessed a remarkable surge in the development and application of Large Language Models (LLMs). These models, such as ChatGPT~\citep{OpenAI2023}, GPT-4~\citep{openai2023gpt4}, and Llama-2~\citep{touvron2023llama}, have demonstrated exceptional performance in various natural language understanding and generation tasks~\citep{zhao2023survey}. The success of LLMs can be attributed to the innovative training techniques employed, including instruction tuning, prompt tuning, Low-Rank Adaptor (LoRA)~\citep{hu2021lora,dettmers2022llmint8}. These advances have made it possible to fine-tune and infer models like Llama-2-7B on consumer-level devices, thereby increasing their accessibility and potential for integration into daily life.

However, despite their impressive capabilities, LLMs are not without limitations. One significant challenge is their susceptibility to variations in input types, which can lead to inconsistencies in output and potentially undermine their reliability in real-world applications. For example, when faced with ambiguous or provocative prompts, LLMs may generate inconsistent or inappropriate responses. To address this issue, several studies have been conducted to assess the robustness of LLM models~\citep{zhu2023promptbench, wang2023chatgpt}. However, these efforts often overlook the importance of re-fine-tuning the models and conducting comprehensive studies of adversarial attacks with known adversarial sample generation mechanisms when full access to the model weights, architecture, and training pipeline is available~\citep{guo-etal-2021-gradient,wallace-etal-2019-universal}.

In this paper, we present an extensive study of three leading open-source LLMs: Llama, OPT, and T5. We evaluate the robustness of various sizes of these models across five distinct NLP classification datasets. To assess their vulnerability to input perturbations, we employ the adversarial geometry attack technique and measure the impact on model accuracy. Furthermore, we investigate the effectiveness of commonly used methods in LLM training, such as LoRA, different precision levels, and variations in model architecture and tuning approaches.

Our work makes several notable contributions to the field of LLM evaluation and robustness:
\begin{enumerate}
    \item We introduce a novel white-box style attack approach that leverages output logits and gradients to expose potential vulnerabilities and assess the robustness of LLMs.
    \item We establish a benchmark for evaluating the robustness of LLMs by focusing on their training strategies, setting the stage for future research in this domain.
    \item Our comprehensive evaluation spans five text classification tasks, providing a broad perspective on the capabilities and limitations of the models across diverse applications.
\end{enumerate}

\section{Related Work}
\subsection{The Evaluation of LLMs}
In recent years, the LLM domain has experienced significant advances~\cite{bommasani2021opportunities, wei2022emergent}.  A large number of exemplary large-scale models such as GPT-4, have emerged, showcasing exceptional performance across various sectors.

Given the remarkable capabilities and broad applications of these models, evaluating their performance has become paramount. Consequently, a significant portion of the research is dedicated to NLP tasks. For instance,~\citep{wang2023chatgpt} examines ChatGPT's performance in sentiment analysis, while~\citep{zhang2023sentiment} offers a comparative analysis with other LLMs. Numerous studies have explored the capabilities of LLM in natural language understanding, including text classification as highlighted by~\citep{liang2022holistic}, and inference as demonstrated by~\citep{qin2023chatgpt}. Additionally, extensive research has been conducted to assess LLMs in generation tasks, encompassing areas such as translation~\citep{wang2023documentlevel}, question answering~\citep{bai2023benchmarking, liang2022holistic}, and summarization~\citep{bang2023multitask}.

Due to the impressive performance of LLMs, there has been widespread attention to their safety and stability. ~\citep{wang2023robustness} conducted an early investigation into ChatGPT and other LLMs and utilized existing benchmarks like AdvGLUE~\citep{wang2022adversarial}, ANLI~\citep{nie-etal-2020-adversarial}, and DDXPlus~\citep{tchango2022ddxplus} for their evaluations.~\citep{zhao2023evaluating} assessed the performance of LLMs in visual inputs and their transferability to other visual-language models. On the topic of adversarial robustness, ~\citep{wang2023decodingtrust} introduced the AdvGLUE++ benchmark and proposed an approach to examine machine ethics through system prompts.~\citep{zhu2023promptbench} proposed a unified benchmark named PromptBench to evaluate the resilience of LLMs against prompts.

However, a significant limitation of the aforementioned studies on robustness is that they focus only on inference-based evaluations. They largely overlook the intricacies of the model's weights and output logits. Furthermore, these studies do not discuss the performance of various LLM techniques. In contrast, in our research, we not only conduct attacks based on the model's parameters and logits, but also actively participate in the model's tuning. Additionally, we place a special emphasis on studying the model size and specific techniques used in LLMs, aspects that previous works have not addressed.

\subsection{Robustness in NLP}

With the rapid advancement in NLP research, its applications have become increasingly prevalent. This ubiquity underscores the growing need for reliable NLP systems that can effectively counteract malign content and misinformation. A seminal work~\citep{jia-liang-2017-adversarial} 
highlighted the vulnerabilities of NLP systems to adversarial attacks.

There are some works about various input perturbations, which could be categorized into three groups: character level, word level, and sentence level~\citep{wei-zou-2019-eda, karimi-etal-2021-aeda-easier,  ma2019nlpaug}.  At the character level, adversarial attacks focus on altering individual characters within a given text. ~\citep{ebrahimi2018hotflip} introduced HotFlip, utilizing gradient information to manipulate characters within text.~\citep{Li_2019} took a different approach by identifying words and modifying their characters. At the word level, adversarial strategies revolve around replacing specific words within the content. For instance,~\citep{alzantot-etal-2018-generating} employs evolutionary algorithms to swap out words with their synonyms. ~\citep{zhang-etal-2019-generating-fluent} utilized probabilistic sampling to generate adversarial examples. Furthermore, some researchers have explored adversarial tactics at the sentence level.~\citep{jia-liang-2017-adversarial} suggested a method that introduces an extraneous sentence to the primary content, aiming to mislead reading models. On the other hand,~\citep{peters-etal-2018-deep} adopted a new approach, where they employed an encoder-decoder framework to rephrase entire sentences. Recently, ~\citep{wang2023decodingtrust} evaluated the robustness and trustworthiness of GPT-3.5 and GPT-4 models, revealing vulnerabilities such as the ease of generating toxic and biased outputs and leaking private information. Despite GPT-4's improved performance on standard benchmarks, it is more susceptible to adversarial prompts, highlighting the need for rigorous trustworthiness guarantees and robust safeguards against new adaptive attacks.

However, the landscape of NLP research is ever-evolving. With the introduction of more sophisticated models boasting novel architectures and training methodologies, there is an growing need to assess the robustness of these newer models. This is especially true for LLMs, which present unique challenges and opportunities in the realm of robustness research.

\section{Preliminaries}
\subsection{Open-source Large Language Models}

We evaluate the following open-source large language models used in our experiments, as shown in Table~\ref{tab:llm}.

\begin{table}[ht!]
    \centering
    \small
    \begin{tabular}{p{0.1\linewidth} p{0.8\linewidth}}
        \toprule
        \textbf{T5} & The Text-to-Text Transfer Transformer (T5) developed by Google Research redefines natural language processing (NLP) tasks by treating them uniformly as text-to-text conversions~\citep{raffel2020exploring}.\\
        \midrule
        \textbf{OPT} & Open Pretrained Transformers (OPT) range from 125M to 175B parameters and are decoder-only models~\citep{zhang2022opt}. These models are trained on a diverse pre-training corpus.  \\
        \midrule
        \textbf{Llama} & Meta AI's Llama is a series of models ranging from 7B to 65B parameters. Llama is trained on a corpus comprising trillions of tokens from publicly available datasets. \\
        \bottomrule
    \end{tabular}
    \caption{The open-source large language models in our study}
    \label{tab:llm}
\end{table}

\subsection{Fine-tuning Techniques}
\label{finetune}
We apply the following fine-tuning techniques in our study.

\paragraph{LoRA} \label{sec:lora}
LoRA innovates in fine-tuning pretrained language models for specific tasks, addressing the inefficiency of full fine-tuning in increasingly large models. By inserting trainable rank decomposition matrices into each layer and freezing the original model weights, LoRA significantly reduces the number of parameters requiring training. 

\paragraph{Quantization}
Quantization in large language models (LLMs) reduces the model size by lowering weight precision, with 8-bit precision presenting challenges due to errors from quantizing large-value vectors. These errors are pronounced in transformer architectures, requiring mixed-precision decomposition. This involves identifying outliers using a threshold, processing them in fp16, and quantizing the rest of the matrix at 8-bit precision. The two parts are then combined. The approach, exemplified by \texttt{LLM.int8()}, aims to make large models more accessible, trading off some performance for significant size reduction.

\paragraph{QLoRA}
Quantized Low-Rank Adapters (QLoRA)~\citep{dettmers2023qlora} introduce an efficient technique to fine-tune large language models by significantly lowering memory requirements. QLoRA combines 4-bit quantization with Low-Rank Adapters, freezing the parameters of a compressed pretrained language model.

\section{Methods}

\begin{figure*}[htbp]
\centering
\includegraphics[width=\linewidth]{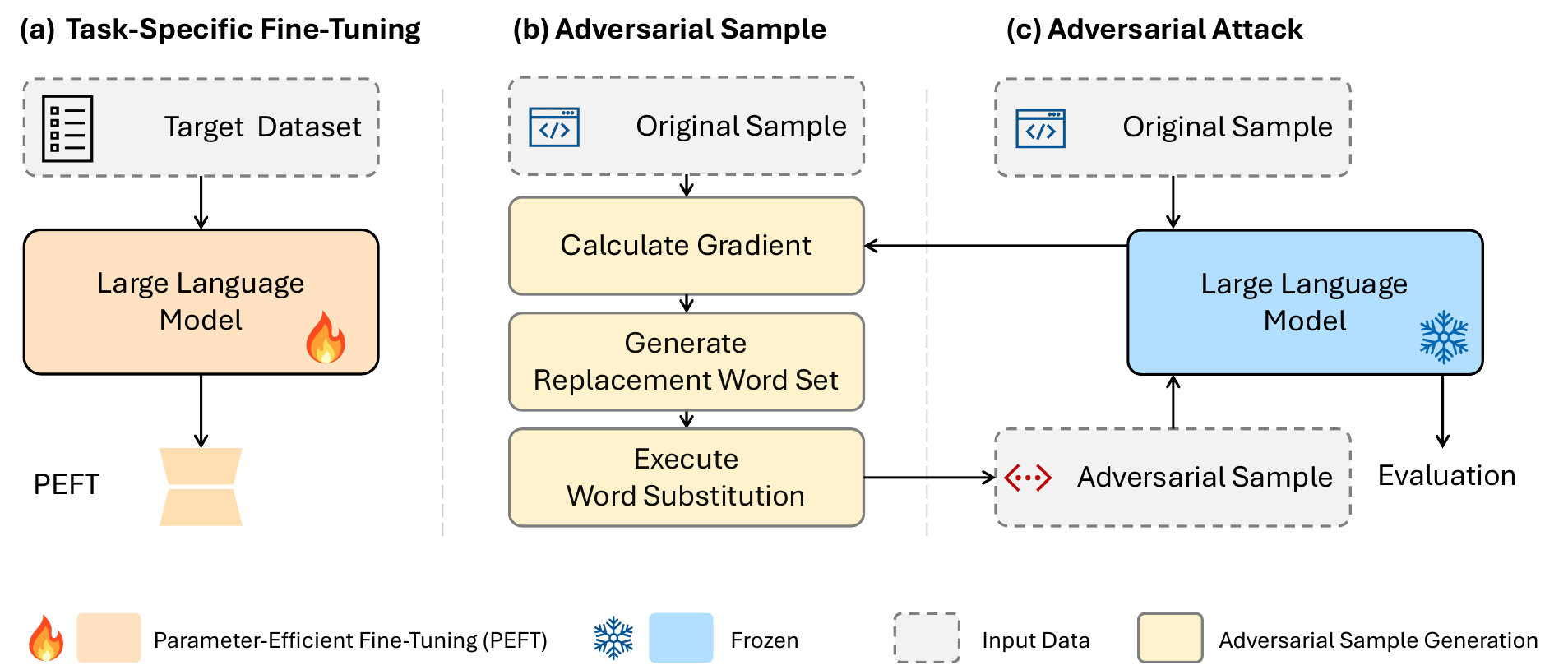}
\caption{The framework of our adversarial robustness assessment }
\label{fig:overview}

\end{figure*}

\subsection{Adversarial Attack}
In this study, our primary concern is text classification. Following the work by ~\citep{meng2022self}, we consider a sample sentence \(S_i = \{w_1, w_2, \ldots, w_L\}\) containing \(L\) words, and its corresponding category label is \(y_i\). Our textual classification system is built upon $n$ LLMs, represented as \(f(\cdot)\), coupled with a prompt indicating the categorization task, denoted as \(P_i\).  In a formal sense:
$\hat{y}_i = f(S_i; P_i)$, where \(\hat{y}_i\) stands for the given answer. The prediction is accurate when \(\hat{y}_i\) equals \(y_i\).

An adversarial attack based on word replacement processes the original sample \(S_i\) to produce an adversarial version \(S_i^{adv}\) by replacing the \(k\)-th word \(w_k\) in \(S_i\) with an alternative word \(w_{adv_k}\). To ensure that the original sample \(S_i\) and its adversarial counterpart \(S_i^{adv}\) maintain semantic similarity, prevalent methodologies typically employ synonymous terms for replacements.

\subsection{Geometry Attack Methodology}
\label{geoattack}

In our research, we extend the basic principles of adversarial attacks in the context of LLMs. Our focus is on exploiting geometric attacks~\citep{meng2020geometry, meng2022self} to assess the vulnerability of LLMs to adversarial perturbations. We propose a systematic methodology grounded in geometric attack insights. The following sections detail the steps of our approach:

\textbf{1) Gradient Computation for Influence Analysis}: We commence by calculating the gradients of the generation loss \(\mathcal{L}_i\) with respect to the embeddings \(e_i\) of input sentence \(\mathit{S}_i\). The cross entropy loss \(\mathcal{L}_i\) measures the dissimilarity between the prediction and label examples in the output space. This computation is essential for all words, including those segmented into sub-tokens. For such words, gradients are computed for each sub-token and subsequently averaged. This initial step is crucial for identifying the words that exert significant influence on \(\mathcal{L}_i\). We determine the gradient of \( \mathcal{L}_i \) with respect to the embedding vector \( e_i \). This step determines the direction in which \( e_i \) should be adjusted to maximize the increase in the loss \( \mathcal{L}_i \). The resulting gradient vector is denoted as \( v_{e_i} = \nabla_{e_i} \mathcal{L}_i \).
    
\textbf{2) Selection of Candidate Words}: Suppose we select a target word \( w_t \) from step 1. Utilizing the DeepFool algorithm~\citep{moosavidezfooli2016deepfool}, we identify potential replacement words, forming a candidate set \( \{w_{t_1}, w_{t_2}, \dots, w_{t_T}\} \). Candidates are filtered based on their cosine similarity to \( w_t \), with those below a defined threshold \( \epsilon \) being excluded. This process ensures that only semantically similar and relevant candidates are considered.

\textbf{3) Optimal Word Replacement and Projection Analysis}: After replacing \( w_t \) with each candidate word, we compute the new text vectors \( \{e_{i_1}, e_{i_2}, \dots, e_{i_T}\} \). For each vector, we define the delta vector \( r_{i_j} \) as \( e_{i_j} - e_i \). The projection of \( r_{i_j} \) onto \( v_{e_i} \) is calculated as \( p_i = r_{i_j} \cdot v_{e_i} \). The optimal replacement candidate \( w_{t_m} \) is selected based on criterion \( m = \arg\max_j \frac{|p_{i_j}|}{||v_{e_i}||} \). This ensures that the chosen word \( w_{t_m} \) induces the largest possible projection \( p_{i_m} \) onto the gradient vector \( v_{e_i} \).

\textbf{4) Iterative Process for Enhanced Adversarial Strength}: The selected word \( w_{t_m} \) replaces \( w_t \) in \( S_i \), updating \( e_i \) to \( e_{i_m} \). This iterative procedure is repeated for \( N \) cycles, where \( N \) is an adjustable parameter in our methodology. Throughout these iterations, an increase in \( \mathcal{L}_i \) should be observed, indicating a continuous enhancement in the adversarial effectiveness of the altered input.

Through this methodically structured process, our research aims to uncover and analyze potential vulnerabilities in LLMs. We refined our methodology to enable prompt fine-tuning for attack generation tasks, expanding its application beyond the previously limited scope of classification tasks.

\section{Experiment Settings}
\subsection{Experiment Pipeline}
This section introduces our methodology for evaluating the robustness of pre-trained LLMs against adversarial attacks. The procedure comprises three principal stages:

\noindent \textbf{1) Model Fine-Tuning}: We fine-tune a pre-trained language model on target dataset with different fine-tuning techniques as described in Sec.~\ref{finetune}, evaluating its accuracy on the corresponding validation set to establish a performance baseline.

\noindent \textbf{2) Adversarial Attack Assessment}: The fine-tuned model undergoes adversarial attacks described in Sec.~\ref{geoattack}, and its performance is assessed on a test dataset altered with adversarial examples.

\noindent \textbf{3) Robustness Evaluation}: We compare the model's accuracy before and after the adversarial attacks to assess its robustness and vulnerability to such manipulations.

\subsection{Datasets}
To evaluate the model's performance under various tasks and its resilience to attacks, we employed five classification datasets, categorized into binary and multi-class classifications. For binary classification, the datasets include IMDB~\citep{maas2011learning}, MRPC~\citep{dolan2005automatically}, and SST-2~\citep{socher2013recursive}, and for multiclass classification, AGNews~\citep{zhang2015character} and DBpedia~\citep{auer2007dbpedia} are used. We will provide a more detailed introduction to these tasks/datasets in Appendix~\ref{sec:datasets_appendix}

\subsection{Evaluation Metrics}

We assess our model's robustness and efficacy using four principal metrics, as described in Table~\ref{tab:metrics}.

\begin{table}[ht!]
    \centering
    \small
    \begin{tabular}{p{0.15\linewidth} p{0.75\linewidth}}
        \toprule
        \textbf{Metrics} & Description\\
        \midrule
        \textbf{Acc} & Accuracy: the model's correct classification rate of untouched input \\
        \midrule
        \textbf{Acc/attack} & Accuracy Under Attack: post-attack classification accuracy revealing adversarial defense\\ 
        \midrule
        \textbf{ASR}& Attack Success Rate: the frequency of accurate predictions turned false by attacks \\ 
        \midrule
        \textbf{Replacement}& Replacement Rate: the extent of input alteration needed to change the model's prediction, indicating sensitivity to perturbations\\
        \bottomrule
    \end{tabular}
    \caption{The evaluation metrics}
    \label{tab:metrics}
\end{table}

\section{Experimental Results}
In this section, we conduct extensive experiments to evaluate the robustness of LLMs across five different datasets. These investigations are guided by three key research questions (RQ):

    \noindent \textbf{RQ1:} How does the robustness of variously sized models differ under adversarial attacks across distinct tasks?
    
    \noindent \textbf{RQ2:} Do contemporary training techniques for LLMs influence their performance and robustness?
    
    \noindent \textbf{RQ3:} How does the model architecture (e.g., fine-tuning with a classification head vs. prompt tuning), affect the robustness of the model?
    
\subsection{Model Size (RQ1)} \label{overall result}
In this section, we analyze the performance metrics of various models across multiple tasks. The datasets under examination include IMDB, SST-2, MRPC, AGNews, and DBPedia. We measure the performance and robustness of LLMs with the metrics Acc, Acc/attack, ASR, and Replacement Rate described in Table~\ref{tab:metrics}. 

The results from the IMDB dataset in Table~\ref{table:imdb} reveal distinct performance variations among different model architectures. In the T5 Series, accuracy generally improves with increasing model size, from 60m to 11b parameters, but the relationship is nonlinear. This suggests that while larger models tend to be more accurate, the accuracy does not increase uniformly with model size. Furthermore, the resilience of these models to adversarial attacks does not follow a simple inverse relationship with model size. The larger T5-11b model shows a more noticeable decrease in accuracy under attack conditions.

For the OPT models, a similar upward trend in accuracy is observed with increasing model size, but the Attack Success Rate (ASR) is lower, suggesting better resistance to attacks. In comparison, the Llama models demonstrate superior performance in both accuracy and robustness against attacks.

\begin{table}[htbp]
\begin{tabular}{ccccc}
\hline
          & Acc    & Acc/attack \(\uparrow\) & ASR \(\downarrow\) & Replacement \\ \hline
T5-60m    & 0.8484 & 0.1256                  & 0.8491             & 0.0929      \\
T5-220m   & 0.8011 & 0.0436                  & 0.9463             & 0.0722      \\
T5-770m   & 0.9048 & 0.1536                  & 0.8312             & 0.1143      \\
T5-3b     & 0.9146 & 0.3259                  & 0.6436             & 0.1413      \\
T5-11b    & 0.9122 & 0.3098                  & 0.6604             & 0.1330      \\ \hline
OPT-125m  & 0.8616 & 0.6637                  & 0.2297             & 0.0365      \\
OPT-350m  & 0.8564 & 0.6924                  & 0.1915             & 0.0305      \\
OPT-1.3b  & 0.9231 & 0.7515                  & 0.1859             & 0.0421      \\
OPT–2.7b  & 0.9198 & 0.7651                  & 0.1682             & 0.0396      \\
OPT-6.7b  & 0.9408 & 0.7864                  & 0.1641             & 0.0528      \\
OPT-13b   & 0.9431 & 0.8016                  & 0.1500             & 0.0671      \\ \hline
Llama-7b  & 0.9483 & 0.8203                  & 0.1350             & 0.0816      \\
Llama-13b & 0.9472 & 0.8237                  & 0.1304             & 0.0875      \\ \hline
\end{tabular}
\caption{IMDB Dataset Results}
\label{table:imdb}
\end{table}

\begin{table}[htbp]
\begin{tabular}{ccccc}
\hline
          & Acc    & Acc/attack \(\uparrow\) & ASR \(\downarrow\) & Replacement \\ \hline
T5-60m    & 0.9083 & 0.2419                  & 0.7304             & 0.1428      \\
T5-220m   & 0.8884 & 0.1228                  & 0.8622             & 0.1611      \\
T5-770m   & 0.8739 & 0.0534                  & 0.9395             & 0.1785      \\
T5-3b     & 0.9495 & 0.1563                  & 0.8437             & 0.1987      \\
T5-11b    & 0.9656 & 0.2248                  & 0.7672             & 0.2043      \\ \hline
OPT-125m  & 0.8807 & 0.7409                  & 0.1587             & 0.0607      \\
OPT-350m  & 0.9011 & 0.7716                  & 0.1437             & 0.0598      \\
OPT-1.3b  & 0.9443 & 0.8227                  & 0.1288             & 0.0733      \\
OPT–2.7b  & 0.8897 & 0.7921                  & 0.1097             & 0.0476      \\
OPT-6.7b  & 0.9693 & 0.8682                  & 0.1043             & 0.0550      \\
OPT-13b   & 0.9656 & 0.7867                  & 0.1853             & 0.0719      \\ \hline
Llama-7b  & 0.9683 & 0.8203                  & 0.1528             & 0.0916      \\
Llama-13b & 0.9632 & 0.8124                  & 0.1566             & 0.0877      \\ \hline
\end{tabular}
\caption{SST-2 Dataset Results}
\label{table:sst2}
\end{table}

\begin{table}[htbp]
\begin{tabular}{ccccc}
\hline
          & Acc    & Acc/attack \(\uparrow\) & ASR \(\downarrow\) & Replacement \\ \hline
T5-60m    & 0.8048 & 0.0325                  & 0.9594             & 0.0720      \\
T5-220m   & 0.8035 & 0.0633                  & 0.9203             & 0.1100      \\
T5-770m   & 0.8924 & 0.1992                  & 0.7771             & 0.1203      \\
T5-3b     & 0.8584 & 0.1074                  & 0.8739             & 0.0917      \\
T5-11b    & 0.8877 & 0.0712                  & 0.9196             & 0.0873      \\ \hline
OPT-125m  & 0.8321 & 0.6504                  & 0.2184             & 0.0332      \\
OPT-350m  & 0.8956 & 0.6741                  & 0.2473             & 0.0437      \\
OPT-1.3b  & 0.9134 & 0.7721                  & 0.1547             & 0.0419      \\
OPT–2.7b  & 0.9128 & 0.7854                  & 0.1396             & 0.0533      \\
OPT-6.7b  & 0.9096 & 0.7902                  & 0.1313             & 0.0579      \\
OPT-13b   & 0.9254 & 0.8183                  & 0.1157             & 0.0560      \\ \hline
Llama-7b  & 0.9277 & 0.8256                  & 0.1101             & 0.0637      \\
Llama-13b & 0.9198 & 0.8107                  & 0.1186             & 0.0742      \\ \hline
\end{tabular}
\caption{MRPC Dataset Results}
\label{table:mrpc}
\end{table}

From Table~\ref{table:sst2} on the SST-2 dataset, there are distinct performance trends. The T5-11b, achieves the highest accuracy of 0.9656. However, its persistence to adversarial attacks is not highest. Notably, the highest ASR within the T5 series is recorded for the T5-770m model, indicating a trade-off as model size increases. In the case of the OPT series, the OPT-6.7b model stands out. However, similar to the IMDB dataset, this model also shows a significant decline in accuracy but more robust than T5 models. One more observation in the OPT series is the overall decrease in ASR with increasing model size, but this trend is disrupted at the 13b parameter mark, where an anomalous increase in ASR is observed. The Llama models, demonstrate consistently high accuracy. It also presents lower ASR compared to T5 models but similar performance to OPT models. For SST-2, the ASR of T5 models exhibit a trend entirely contrary to that observed for MRPC. It reaches its minimum at the T5-770m model. For the OPT models, although their ASR is much lower compared to the T5 series, there is a consistent decrease in ASR as the size of the OPT models increases. Regarding the Llama models, the 7b model slightly outperforms the 13b in terms of accuracy and ASR. 

\begin{table}[htbp]
\begin{tabular}{ccccc}
\hline
          & Acc    & Acc/attack \(\uparrow\) & ASR \(\downarrow\) & Replacement \\ \hline
T5-60m    & 0.8606 & 0.3608                  & 0.5807             & 0.1740      \\
T5-220m   & 0.9084 & 0.5370                  & 0.4098             & 0.1864      \\
T5-770m   & 0.9278 & 0.6597                  & 0.2896             & 0.1860      \\
T5-3b     & 0.9193 & 0.7267                  & 0.2102             & 0.1834      \\
T5-11b    & 0.9212 & 0.8469                  & 0.1531             & 0.1867      \\ \hline
OPT-125m  & 0.8152 & 0.6040                  & 0.2591             & 0.0809      \\
OPT-350m  & 0.8321 & 0.6036                  & 0.2746             & 0.0864      \\
OPT-1.3b  & 0.8806 & 0.6316                  & 0.2828             & 0.0912      \\
OPT–2.7b  & 0.9175 & 0.7028                  & 0.2340             & 0.0833      \\
OPT-6.7b  & 0.9341 & 0.7143                  & 0.2353             & 0.0756      \\
OPT-13b   & 0.9456 & 0.7745                  & 0.1809             & 0.0941      \\ \hline
Llama-7b  & 0.9328 & 0.7315                  & 0.2158             & 0.0864      \\
Llama-13b & 0.9338 & 0.7688                  & 0.1767             & 0.0837      \\ \hline
\end{tabular}
\caption{AGNews Dataset Results}
\label{table:agnews}
\end{table}

\begin{table}[htbp]
\begin{tabular}{ccccc}
\hline
          & Acc    & Acc/attack \(\uparrow\) & ASR \(\downarrow\) & Replacement \\ \hline
T5-60m    & 0.9817 & 0.3974                  & 0.5952             & 0.1187      \\
T5-220m   & 0.9765 & 0.4082                  & 0.5819             & 0.1234      \\
T5-770m   & 0.9921 & 0.7476                  & 0.2464             & 0.1483      \\
T5-3b     & 0.9914 & 0.8608                  & 0.1317             & 0.1550      \\
T5-11b    & 0.9919 & 0.8815                  & 0.1113             & 0.1724      \\ \hline
OPT-125m  & 0.9034 & 0.6512                  & 0.2792             & 0.0534      \\
OPT-350m  & 0.9511 & 0.6718                  & 0.2937             & 0.0305      \\
OPT-1.3b  & 0.9784 & 0.7046                  & 0.2798             & 0.0496      \\
OPT–2.7b  & 0.9822 & 0.7513                  & 0.2351             & 0.0552      \\
OPT-6.7b  & 0.9907 & 0.7780                  & 0.2147             & 0.0641      \\
OPT-13b   & 0.9916 & 0.7912                  & 0.2021             & 0.0576      \\ \hline
Llama-7b  & 0.9908 & 0.7596                  & 0.2333             & 0.0881      \\
Llama-13b & 0.9921 & 0.7286                  & 0.2656             & 0.0912      \\ \hline
\end{tabular}
\caption{DBPedia Dataset Results}
\label{table:dbpedia}
\end{table}

In Tables~\ref{table:agnews} and Tables ~\ref{table:dbpedia}, analyzing results from multi-class classification tasks, a distinct pattern emerges. These datasets reveal enhanced stability against synonym substitution attacks. For T5 models, the data shows a lower ASR on these tasks compared to binary datasets, suggesting a better resistance to attacks in complex classification scenarios. In contrast, OPT and Llama models exhibit a higher ASR on the AGNews and DBpedia14 datasets. Another result is that for both T5 and OPT series, there is a marked decline in ASR around the 770 million or 1 billion parameter threshold. This indicates an increased robustness and better handling of adversarial attacks with the scale-up of model size. 

\subsubsection{Analysis}
When examining the accuracy of the model, we observed a trend where the accuracy gradually increases with the growth in model size. However, after reaching a certain size threshold, the accuracy tends to saturate, stabilizing around specific values. This phenomenon is particularly pronounced when tested on datasets like DBpedia. When comparing different models operating at the same parameter scale, their performances were found to be quite similar, without any significant disparities.

\begin{figure*}[htbp]
\centering
\begin{subfigure}{.3\textwidth}
\includegraphics[width=\linewidth]{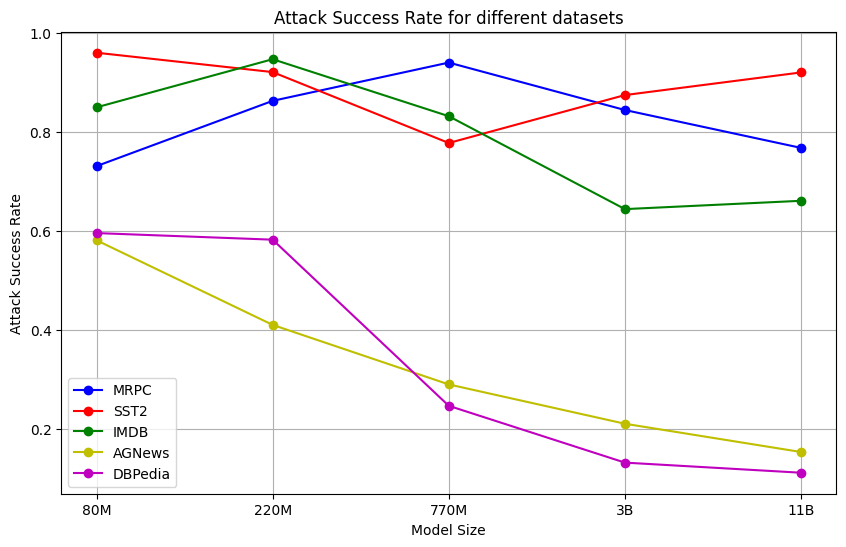}
\caption{T5 model}
\label{fig:t5}

\end{subfigure}\hfill 
\begin{subfigure}{.3\textwidth}
\includegraphics[width=\linewidth]{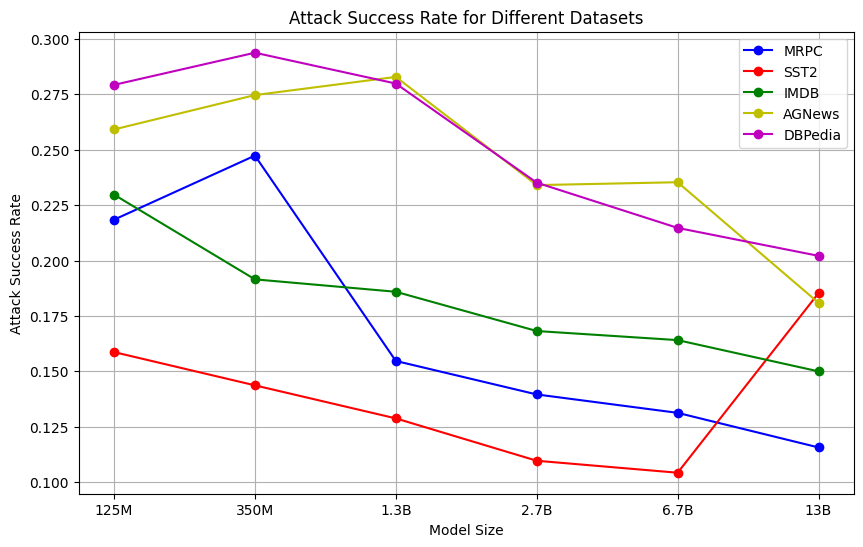}
\caption{OPT model}
\label{fig:opt}

\end{subfigure}\hfill
\begin{subfigure}{.3\textwidth}
\includegraphics[width=\linewidth]{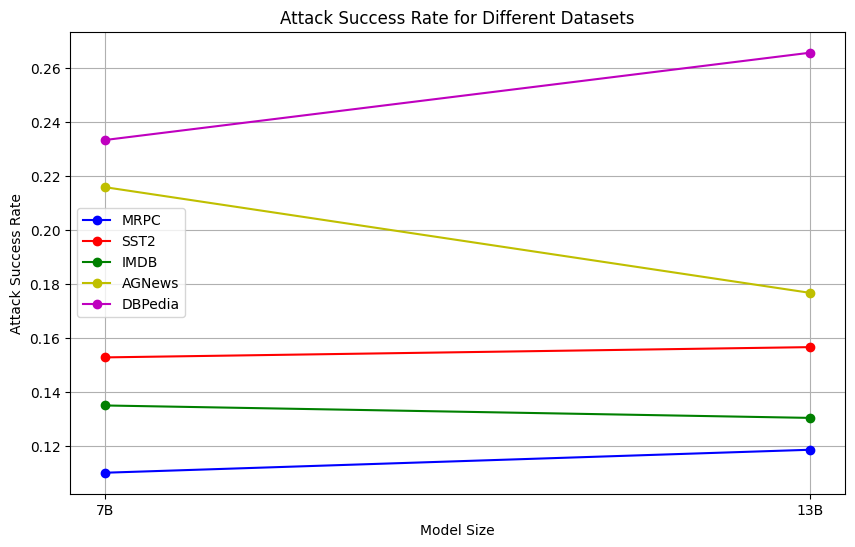}
\caption{Llama model}

\label{fig:Llama}
\end{subfigure}
\caption{The experimental results of different models on various datasets.}

\label{fig:models}
\end{figure*}

However, the experiments related to robustness revealed more distinct differences. From Figure ~\ref{fig:t5}, Figure~\ref{fig:opt} and Figure~\ref{fig:Llama}, we have more intuitive results. Observing the performance of a uniform model across various datasets, we made the following observations:

\noindent \textbf{T5 Model}: As the size of the T5 model increases, its ASR gradually decreases. This suggests that larger models, with more parameters, tend to have a deeper understanding of language. As a result, they can maintain stronger stability in the face of various disturbances. However, on datasets like MRPC and SST-2, there were noticeable fluctuations in performance. One possible explanation for this is that as the model size grows, the words selected based on the model's gradient become more precise and have a more significant impact on the results. This introduces a trade-off related to model size.

\noindent \textbf{OPT Model}: For the OPT model, a similar trend was observed across most datasets. As the model size increased, its robustness generally improved, aligning with the observations made for the T5 model.

\noindent \textbf{Llama Model}: For the Llama model, the differences in performance between the two sizes were minimal. This suggests that the size variation did not significantly influence the model's robustness.

However, when comparing different models, the disparities become even more pronounced. It is obvious that the T5 model's ASR and replacement rate are significantly higher than those of OPT and Llama. This indicated that Decoder-only Causal LMs have higher robustness against encoder-decoder architectures under synonym substitution adversarial attacks.

\subsection{LLMs Fine-tuning Techniques (RQ2)}
\subsubsection{Instruction Tuning}
To study the impact of instruction tuning on model robustness, we compared the performance of Flan-T5 with the standard T5. The Flan-T5 is an advanced variant of T5 that has undergone instruction tuning across over a thousand downstream tasks. In contrast, the traditional T5 was not trained with such an extensive procedure.

Based on our experimental results, as shown in the table, there is a significant decline in accuracy for both T5 and Flan-T5 under adversarial attacks. This observation indicates that models, irrespective of whether they have undergone instruction tuning, remain susceptible to adversarial manipulations. Furthermore, consistent with our previous findings, we noticed that as the model size increases, the attack success rate tends to decline.

\begin{figure}[htbp]
\includegraphics[width=0.6\linewidth]{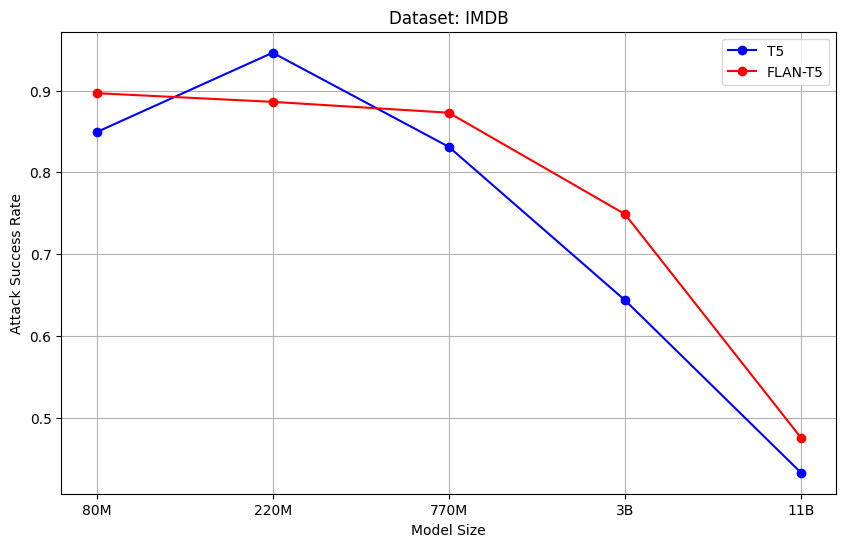}
\caption{The experimental results of T5 and Flan-T5 on IMDB dataset}
\label{fig:flan}
\end{figure}

Interestingly, as shown in Fig \ref{fig:flan}, our results indicated that Flan-T5 exhibits a higher ASR than the standard T5. This suggests that models subjected to instruction tuning, like Flan-T5, can be more easily compromised. We hypothesize the primary reason for this observation:

The instruction tuning process for Flan-T5 encompassed datasets similar to IMDB. This might have rendered the model with a deeper understanding of tasks related to this data. As a result, attackers could more easily pinpoint words in the input that were influential and susceptible to replacement.
\begin{table}[htbp]
\small
\centering 
\label{instruction}
\scalebox{0.74}{
\begin{tabular}{@{}ccccc@{}}
\toprule
             & Acc    & Acc/attack \(\uparrow\) & ASR \(\downarrow\)   & Replacement \\ \midrule
T5-60m       & 0.8484 & 0.1256     & 0.8491 & 0.0929      \\
Flan-T5-60m  & 0.8453 & 0.0882     & 0.8968 & 0.0820       \\
T5-220m      & 0.8011 & 0.0436     & 0.9463 & 0.0722      \\
Flan-T5-220m & 0.8777 & 0.0996     & 0.8862 & 0.0978      \\
T5-770m      & 0.9048 & 0.1536     & 0.8312 & 0.1143      \\
Flan-T5-770m & 0.9141 & 0.1171     & 0.8729 & 0.1106      \\
T5-3b        & 0.9146 & 0.3259     & 0.6436 & 0.1413      \\
Flan-T5-3b   & 0.9228 & 0.2328     & 0.7489 & 0.1261      \\
T5-11b       & 0.9348 & 0.4904     & 0.4752 & 0.1326      \\
Flan-T5-11b  & 0.9122 & 0.3098     & 0.6604 & 0.1330       \\ \bottomrule
\end{tabular}
}
\caption{Experimental results comparing T5 and Flan-T5 after instruction tuning using the IMDB dataset.}
\end{table}
\hfill 



\begin{figure}[htbp]
    \centering
    \begin{minipage}[b]{0.35\textwidth}
        \centering
        \includegraphics[width=\textwidth]{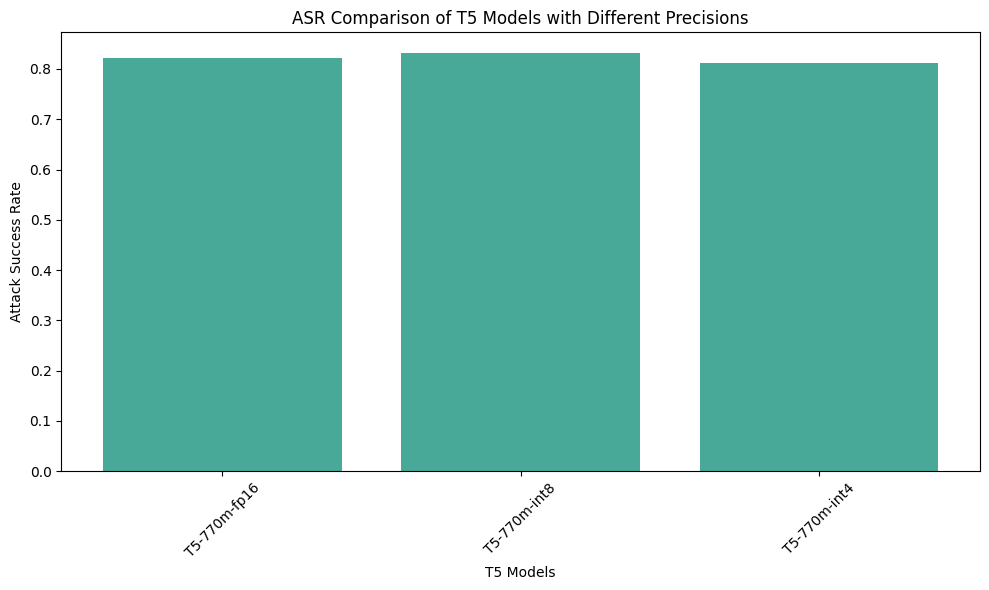}
        \caption{Different precisions on the IMDB dataset (T5 model)}
        \label{fig:t5-precisions}
    \end{minipage}
    \hfill
    \begin{minipage}[b]{0.35\textwidth}
        \centering
        \includegraphics[width=\textwidth]{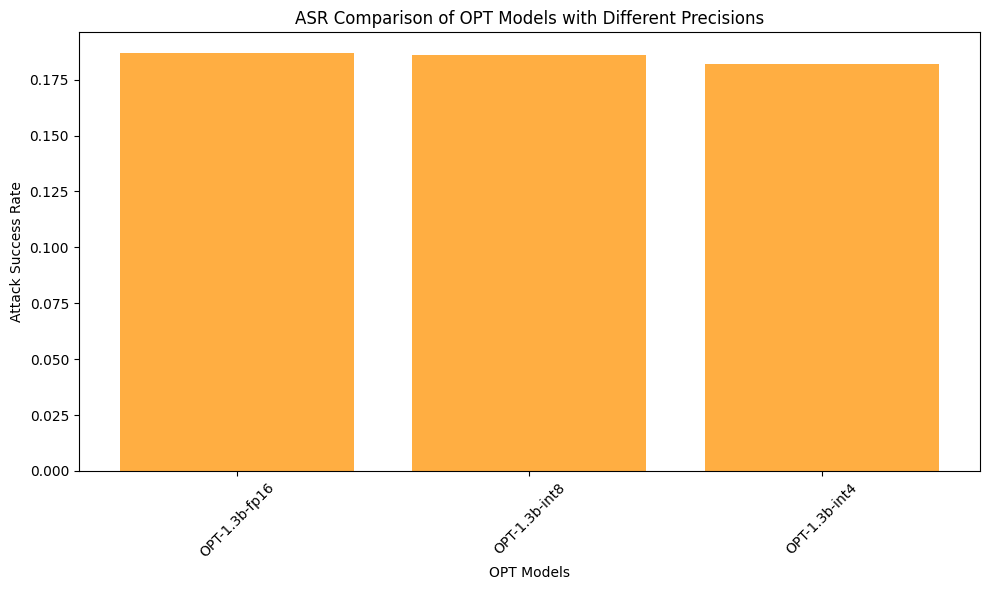}
        \caption{Different precisions on the IMDB dataset (OPT Model)}
        \label{fig:precisions}
    \end{minipage}
    \hfill
    \begin{minipage}[b]{0.35\textwidth}
        \centering
        \includegraphics[width=\textwidth]{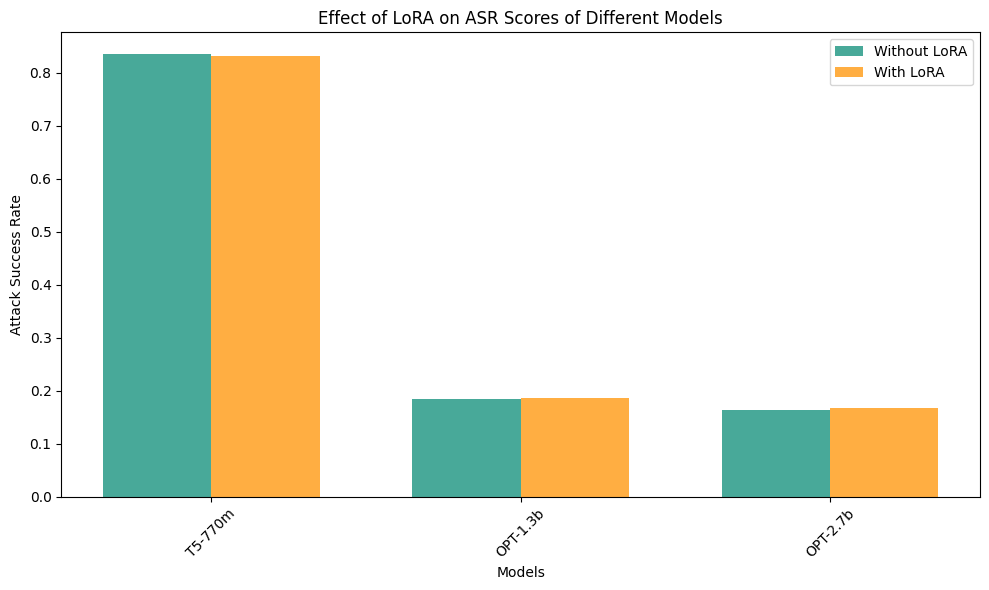}
        \caption{Results of LoRA on IMDB dataset}
        \label{fig:lora}
    \end{minipage}
    \caption{Comparison of different models and precisions on the IMDB dataset}
\end{figure}

\subsubsection{Precisions}

In machine learning, balancing model size with precision is crucial. Model size indicates capacity, and precision affects information granularity. Larger models typically perform better but require more computational resources. Techniques like quantization and precision adjustments help deploy these models more efficiently. We studied the impact of precision settings on the robustness of T5-770m and OPT-1.3b models by comparing their performance under various precisions.

For the T5-770m and OPT-1.3b models, it's clear that as precision changes from fp16 to int4, there isn't a significant drop in their inherent accuracy. This indicates that models can handle reduced precision without compromising their general performance drastically. What's more, across different precision settings, the attack success rate for the T5-770m models remains fairly higher, which shows the same conclusion as in \ref{overall result}. However, the precision settings do not show a consistent pattern of influence on the ASR and replacement rate.

In essence, while different models exhibit different robustness against adversarial attacks, the precision settings do not play a significant role in this robustness.

\begin{table}[htbp]
\small
\centering 
\label{Precisions}
\scalebox{0.75}{
\begin{tabular}{@{}ccccc@{}}
\toprule
              & Acc    & Acc/attack \(\uparrow\) & ASR \(\downarrow\)   & Replacement \\ \midrule
T5-770m-fp16  & 0.9106 & 0.1631     & 0.8208 & 0.1196      \\
T5-770m-int8  & 0.9048 & 0.1536     & 0.8312 & 0.1143      \\
T5-770m-int4  & 0.9210 & 0.1725     & 0.8127 & 0.1211      \\
OPT-1.3b-fp16 & 0.9218 & 0.7496     & 0.1868 & 0.0536      \\
OPT-1.3b-int8 & 0.9231 & 0.7515     & 0.1859 & 0.0421      \\
OPT-1.3b-int4 & 0.9207 & 0.7531     & 0.1820 & 0.0498      \\ \bottomrule
\end{tabular}}
\caption{Results of T5-770m and OPT-1.3b models under different precision settings, including fp16, int8 and int4. The performance is evaluated with IMDB dataset.}

\end{table}

\begin{table}[htbp]
\small
\centering 
\label{LoRA}
\scalebox{0.75}{
\begin{tabular}{@{}ccccc@{}}
\toprule
              & Acc    & Acc/attack \(\uparrow\) & ASR \(\downarrow\)   & Replacement \\ \midrule
T5-770m       & 0.9067 & 0.1499     & 0.8347 & 0.1036      \\
T5-770m-Lora  & 0.9048 & 0.1536     & 0.8312 & 0.1143      \\
OPT-1.3b      & 0.9135 & 0.7448     & 0.1847 & 0.0366      \\
OPT-1.3b-LoRA & 0.9231 & 0.7515     & 0.1859 & 0.0421      \\
OPT-2.7b      & 0.9266 & 0.7741     & 0.1646 & 0.0452      \\
OPT-2.7b-LoRA & 0.9198 & 0.7651     & 0.1682 & 0.0396      \\ \bottomrule
\end{tabular}}
\caption{Results of the T5-770m, OPT-1.3b, and OPT-2.7b models' performance with and without the application of LoRA, using the IMDB dataset.}

\end{table}

\subsubsection{LoRA}

As mention in Sec \ref{sec:lora}, LoRA has been a groundbreaking approach, bringing about significant reductions in memory requirements during model training. In this case, the potential trade-off in question is model robustness.

For our investigation, we selected the T5-770m, OPT-1.3b, and OPT-2.7b models. Experiments were conducted under two conditions for each model: with and without the application of LoRA. The IMDB dataset served as our benchmark for this analysis.

The experiments show that adversarial attacks significantly reduce accuracy across all models, regardless of LoRA's use. However, crucially, both the attack success rate and replacement rate, key measures of resilience against adversarial tactics, were unaffected by LoRA. This indicates that while LoRA enhances optimization, it doesn't negatively impact the model's defense against adversarial attacks, providing optimization benefits without sacrificing robustness. 

\subsection{Model Architectures (RQ3)}

The architecture of a model's output space significantly influences its performance and resilience against adversarial attacks. For models with a classification head, the output is simplified to a binary decision, contrasting with OPT models without such a head, which must identify 'negative' or 'positive' labels from a vast vocabulary. This distinction impacts the model's accuracy.

Our data shows that smaller models with a classification head are more accurate than headless ones due to their simplified output space, which aids decision-making, especially in models with limited processing power. Moreover, models with a head reach their peak performance faster, achieving accuracy saturation more quickly.

However, an intriguing observation is the heightened attack success rate for models with a classification head. On the surface, this suggests that launching adversarial attacks against these models is a more straightforward task.One main factor contributes to this vulnerability is DeepFool's efficacy with last layer FFN: In such models, DeepFool can more readily discern the optimal direction for launching its attack, amplifying the ASR. This marked efficiency underscores a reduced robustness in these models against adversarial intrusions.

\begin{table}[htbp]
\small
\centering 
\label{Classification Head}
\scalebox{0.9}{
\begin{tabular}{@{}ccccc@{}}
\toprule
              & Acc    & Acc/attack \(\uparrow\) & ASR \(\downarrow\)   & Replacement \\ \midrule
OPT-125m      & 0.8616 & 0.6637     & 0.2297 & 0.0365      \\
OPT-125m-head & 0.9074 & 0.6215     & 0.3151 & 0.0476      \\
OPT-350m      & 0.8564 & 0.6924     & 0.1915 & 0.0305      \\
OPT-350m-head & 0.9152 & 0.6643     & 0.2741 & 0.0682      \\
OPT-1.3b      & 0.9231 & 0.7515     & 0.1859 & 0.0421      \\
OPT-1.3b-head & 0.9316 & 0.7621     & 0.1819 & 0.0533      \\
OPT–2.7b      & 0.9198 & 0.7651     & 0.1682 & 0.0396      \\
OPT–2.7b-head & 0.9367 & 0.7704     & 0.1775 & 0.0516      \\
OPT-6.7b      & 0.9408 & 0.7864     & 0.1641 & 0.0528      \\
OPT-6.7b-head & 0.9422 & 0.7765     & 0.1759 & 0.0627      \\
OPT-13b       & 0.9431 & 0.8016     & 0.1500 & 0.0671      \\
OPT-13b-head  & 0.9427 & 0.7877     & 0.1644 & 0.0641      \\ \bottomrule
\end{tabular}
}
\caption{Experimental results of OPT models with/without classification head with IMDB dataset.}

\end{table}

\begin{figure}[htbp]
\centering
\includegraphics[width=0.6\linewidth]{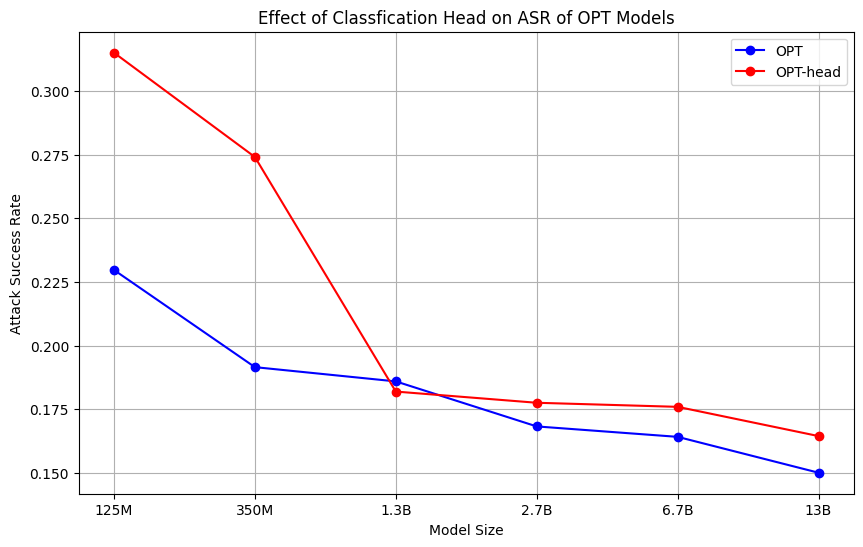}

\caption{Results of classfication head on IMDB dataset}
\label{fig:head}
\end{figure}

\subsection{Attack Examples}

In this section, we are going to show some adversarial examples for different tasks and models in Table ~\ref{tab:adversarial_examples} 

\begin{table*}[htbp]
\centering
\caption{Adversarial Examples for Various Tasks and Models}
\begin{tabularx}{\linewidth}{X}
\toprule
\textbf{Task Description and Adversarial Example} \\
\midrule
\textit{Dataset:} IMDB \\
\textit{Model:} T5-780M \\
\textbf{Original:} "Choose the sentiment of this review?  Expensive \textcolor{blue} {lunch} meals. Fried \textcolor{blue}{pickles} were good. Waitress \textcolor{blue}{messed} up 2 orders out of 4. "Dont" \textcolor{blue}{think} "Ill" return. Asked for no cheese \textcolor{blue}{waitress} \textcolor{blue}{joked} extra cheese then brought my meal with cheese. Better \textcolor{blue}{places} to \textcolor{blue}{eat} in area. " \\
\textbf{Adversarial:} "Choose the sentiment of this review?  Expensive \textcolor{red}{brunch} meals. Fried \textcolor{red}{marinated} were good. Waitress \textcolor{red}{eventhough} up 2 orders out of 4. "Dont" \textcolor{red}{imagining} "Ill" return. Asked for no cheese \textcolor{red}{miss jokingly} extra cheese then brought my meal with cheese. Better \textcolor{red}{place} to \textcolor{red}{devoured} in area." \\
\midrule
\textit{Dataset:} MRPC \\
\textit{Model:} OPT-2.7B \\
\textbf{Original:} "Do these two sentences mean the same thing? Crews \textcolor{blue}{worked} to \textcolor{blue}{install} a \textcolor{blue}{new culvert} and prepare the \textcolor{blue}{highway} so \textcolor{blue}{motorists} could use the eastbound lanes for \textcolor{blue}{travel} as storm \textcolor{blue}{clouds threatened} to \textcolor{blue}{dump} more \textcolor{blue}{rain}. Crews \textcolor{blue}{worked} to install a  \textcolor{blue}{new culvert} and repave the highway so motorists could use the eastbound lanes for \textcolor{blue}{travel}." \\
\textbf{Adversarial:} "Do these two sentences mean the same thing? Crews \textcolor{red}{acted} to \textcolor{red}{mount} a \textcolor{red}{nouvelle drains} and prepare the \textcolor{red}{avenue} so \textcolor{red}{chauffeurs} could use the eastbound \textcolor{red}{routing} for \textcolor{red}{voyage} as stormy \textcolor{red}{haze threatens} to \textcolor{red}{spill} more \textcolor{red}{rainfall}. Crews worked to install a \textcolor{red}{newer septic} and repave the highway so motorists could use the eastbound lanes for \textcolor{red}{tours}" \\
\midrule
\textit{Dataset:} AGNews \\
\textit{Model:} Llama-7B \\
\textbf{Original:} "Dial M for \textcolor{blue}{Music} Mobile-phone \textcolor{blue}{makers} scored a surprising hit four years ago when they \textcolor{blue}{introduced} handsets equipped with tiny digital cameras. Today, nearly one-third of the cell phones sold worldwide do double duty as cameras." \\
\textbf{Adversarial:} "Dial M for \textcolor{red}{Melody} Mobile-phone \textcolor{red}{manufacturers} scored a surprising hit four years ago when they \textcolor{red}{unveiled} handsets equipped with tiny digital cameras. Today, nearly one-third of the cell phones sold worldwide do double duty as cameras." \\
\bottomrule
\end{tabularx}
\label{tab:adversarial_examples}
\end{table*}

\section{Conclusion}

This paper utilized a novel geometric adversarial attack method to assess the robustness of leading LLMs, utilizing advanced fine-tuning techniques for task-specific model adaptation. Our groundbreaking approach revealed that these models exhibit variable sensitivity to adversarial attacks, influenced by their size and architectural differences. This indicates inherent vulnerabilities in LLMs, yet suggests potential resilience in certain configurations. Contrary to expectations, LLM-specific techniques did not markedly reduce robustness. Future research could explore models like RLHF and model parallelism approaches within this framework. Additionally, the evolution of more complex adversarial attacks promises deeper insights into LLM strengths and weaknesses.

\section*{Ethics statement}
In our research, we employ adversarial attack methodologies to generate text, aiming to evaluate the robustness of LLMs against inputs. However, we acknowledge the ethical implications associated with the use of adversarial attacks. One primary concern is the potential generation of harmful information. This includes text that may be offensive, misleading, or harmful in other ways. Therefore, one should be cautious when taking such methods into practical use.



\bibliographystyle{ACM-Reference-Format}
\bibliography{sample-base}

\appendix

\section{Research Methods}
\label{sec:appendix}
\subsection{Experiments Setup} \label{sec: experiment settings}
In our study, we employ pretrained weights from HuggingFace and use int8 quantization for GPU memory optimization. We also standardize the use of LoRA to reduce training parameters. For models under 3 billion parameters, experiments are conducted on an NVIDIA RTX 3090 (24GB), whereas models above 3 billion parameters are tested on NVIDIA RTX A6000 (48GB) or NVIDIA A100 (40GB), catering to both fine-tuning and attack simulations.

\subsection{Dataset} \label{sec:datasets_appendix}
In this section, we will provide more details about the datasets used in this work.

\textbf{IMDB}: This dataset contains 50,000 movie reviews for sentiment analysis, equally divided between positive and negative sentiments.

\textbf{SST-2}: An extension of the original SST, it focuses on the binary classification of sentiments in movie review sentences.

\textbf{MRPC}: A corpus for paraphrase identification, it includes sentence pairs from online news sources, annotated for semantic equivalence.

\textbf{AGNews}: This news categorization dataset comprises 120,000 training and 7,600 test samples across four categories: World, Sports, Business, and Science/Technology.

\textbf{DBpedia}: A large-scale, multi-class dataset from the DBpedia knowledge base, featuring 560,000 training and 70,000 test samples across 14 categories.

The statistics of these five datasets are presented in Table \ref{dataset}.

\begin{table}[hb]
\centering
\caption{Statistics of the Datasets}
\label{dataset}
\begin{tabular}{@{}ccccc@{}}
\toprule
Dataset & Labels & Avg Length & Train & Test \\ 
\midrule
IMDB & 2 & 279.48 & 25000 & 25000 \\
MRPC & 2 & 52.89 & 3670 & 1730 \\ 
SST-2 & 2 & 19.81 & 67300 & 1820 \\ 
AGNews & 4 & 43.93 & 120000 & 7600 \\
DBPedia & 14 & 55.14 & 560000 & 7000 \\ 
\bottomrule
\end{tabular}
\end{table}

\end{document}